\documentclass[letterpaper]{article} 
\usepackage[preprint]{aaai2027}  
\usepackage[hyphens]{url}  
\usepackage{graphicx} 
\usepackage{amsmath}
\usepackage{amssymb}
\usepackage{algorithm}
\usepackage{algorithmic}
\usepackage{booktabs}
\usepackage{xcolor} 
\definecolor{profrev}{RGB}{140,70,20} 
\usepackage{natbib}  
\usepackage{caption} 
\newcommand{\cmin}{C_{\min}}

\title{Arm2Air: Cross-Embodiment Skeleton Transfer for 3D Relay Formation}

\author{
    Dohun Lee\equalcontrib,
    Kyeonghyun Yoo\equalcontrib,
    Seokmin Kim,\\
    Byongho Lee,
    Seungjoo Oh,
    Hwangnam Kim\corresponding
}

\affiliations{
    Department of Electrical and Electronic Engineering,
    Korea University, Seoul, Republic of Korea\\
    \{luke0911, seven1705, slartler17, unlike96, seungjoooh, hnkim\}@korea.ac.kr
}

\begin{document}

\maketitle

\begin{abstract}
Unmanned aerial vehicle (UAV) relay networks can restore connectivity after communication infrastructure is damaged. Urban relay placement is difficult because line-of-sight blockage, communication range, altitude, and three-dimensional obstacles must be considered jointly. Arm2Air transfers obstacle-avoidance skeletons from robot arms to UAV relay placement through cross-embodiment transfer. Source-domain robot-arm motions from a pretrained Neural MP model are converted into ordered skeletons that pretrain a transformer-based transfer platform, which is then adapted to the UAV domain using limited target data and Low-Rank Adaptation. The transferred skeleton initializes a relay chain that is refined for connectivity, bottleneck capacity, delay, and movement cost. On nine held-out high-clutter 3D urban maps, Arm2Air reduced median end-to-end planning runtime by 64.9 percent relative to the fastest conventional planner. On the high-obstruction group of a separate 30-map dense urban holdout, it increased bottleneck capacity by 32.6 percent, reduced capacity variance by 74.7 percent, reduced maximum hop distance by 13.2 percent, reduced hop-distance variance by 75.2 percent, and reduced relay displacement by 16.9 percent relative to IMPC-MD. With only three target-domain training maps, Arm2Air reduced relay-position root mean square error by 53.6 percent relative to training from scratch while updating 0.134 million parameters, compared with 1.383 million for Scratch and Full Fine-tuning. These results demonstrate computationally and data-efficient UAV relay placement and suggest a broader principle for transferring ordered structural priors across heterogeneous embodied tasks.
\end{abstract}

\section{Introduction}

Unmanned aerial vehicle (UAV) relay networks offer rapidly deployable aerial backbones for disaster areas, temporary infrastructure, and damaged networks \citep{Zeng2016, Mozaffari2019, peng2026resilient}. Unlike a general UAV network, where nodes independently cover an area or collect data, a relay backbone forms a serial communication chain. A mission UAV travels to a distant operating region to perform tasks such as mapping, search-and-rescue, or surveillance, while intermediate relay UAVs position themselves to maintain a communication link with the ground control station (GCS), forwarding mission data and relaying control commands. This structure resembles a serial robotic manipulator: just as an end-effector reaches a target while the remaining joints reconfigure around obstacles to support the motion, relay UAVs must reposition under connectivity, collision-avoidance, and communication constraints to preserve a stable end-to-end communication chain. Such networks must jointly consider 3D placement and mobility, a line-of-sight (LoS)-dominant channel, energy, and quality-of-service (QoS) constraints \citep{Zeng2019, Wu2021, joo2026distributed, 10798419}. Forming a stable backbone in an urban environment is therefore not a shortest-path problem. Each relay must maintain LoS to its neighbors, remain within a maximum range, avoid obstacles, maintain safe separation, and stay within a low-altitude corridor, while backbone performance is limited by the bottleneck capacity of the weakest hop and the accumulated end-to-end delay.

This difficulty stems from the complexity of 3D UAV planning. Generalized obstacle-avoidance motion planning is fundamentally hard \citep{Reif1979}. UAVs must also handle nonlinear dynamics, disturbances, and state uncertainty \citep{Goerzen2010}. Sampling-based planners such as RRT and PRM further depend heavily on the sample count and environment complexity \citep{Karaman2011, Garibeh2025, saeed2022optimal, 4082128}. Communication constraints make the problem harder still. A collision-free path does not guarantee a good backbone: blocked LoS or a weak hop can collapse the network. Joint trajectory and power optimization in multi-UAV relays is also non-convex \citep{Zhang2018Relay, wan2022accurate, du2023multi, wang2024gnn, 777092.777167}.

On larger urban maps, direct 3D planning incurs up to two orders of magnitude higher computational cost than its 2D counterpart, as quantified in our experiments (see Fig.~\ref{fig:arm2air_2d3d}). This scaling behavior motivates a transfer-based initialization that avoids repeated direct 3D search. To avoid this cost, we exploit structural knowledge learned on a different platform: the serial robot arm, which, despite its different dynamics and control, shares the same ordered start-to-goal chain structure as a UAV relay backbone.

We propose Arm2Air, which transfers obstacle-avoidance geometric skeletons from heterogeneous robot arms to UAV relay backbone formation. Arm2Air does not transfer actuator-level variables such as joint angles, torques, or low-level control commands. Instead, a pretrained Neural MP model~\citep{dalal2025neural} generates source-domain robot-arm motions, which are converted through forward kinematics into ordered geometric skeletons. These skeletons pretrain a transformer-based transfer platform, and low-rank adaptation (LoRA) adapts the pretrained platform to the UAV domain. At inference time, a source skeleton is aligned with the gateway-to-target axis to initialize the backbone, the skeleton-conditioned transfer platform proposes target-domain relay coordinates, and a refinement stage over LoS, hop distance, bottleneck capacity, delay, and movement cost produces the final backbone. This view is consistent with prior multi-robot transfer across differing kinematics, link lengths, and degrees of freedom \citep{Devin2017, Chen2018, che2026lora, zhuang2020comprehensive, weiss2016survey, augustin2016study}.

The main contributions are as follows.
\begin{itemize}
    \item We formulate urban UAV relay backbone formation as a combined problem that couples 3D path planning and communication constraints.
    \item Rather than relying on direct 3D replanning alone, we introduce a representation-level approach that transfers ordered obstacle-aware structures across heterogeneous embodiments.
    \item We propose a skeleton-conditioned, transformer-based transfer platform with LoRA adaptation and communication-aware refinement for efficient 3D relay formation.
\end{itemize}

The remainder of this paper is organized as follows. The Related Work section reviews 3D UAV planning, relay-network optimization, and cross-embodiment learning. The System Design section formulates the relay placement problem and presents skeleton transfer, LoRA adaptation, and communication-aware refinement. The Experiments section describes the evaluation protocol and reports planning, communication, movement, transfer-efficiency, and ablation results. The Conclusion summarizes the findings, limitations, and future directions.

\section{Related Work}
\label{sec:related_work}

\subsection{UAV Motion Planning in 3D Environments}

UAV motion planning is more complex than ordinary ground-robot path planning because it must jointly account for a 3D operating space, nonlinear dynamics, disturbances, state uncertainty, and obstacle avoidance. Generalized obstacle-avoidance motion planning is fundamentally hard in terms of computational complexity \citep{Reif1979}, and work from a UAV-guidance perspective likewise emphasizes that UAVs must consider 3D environments and dynamic constraints together \citep{Goerzen2010}. Sampling-based planners of the RRT and PRM families are widely used for high-dimensional path planning, but their solution quality and cost depend heavily on the number of samples, the obstacle distribution, and the cost of collision checking \citep{Karaman2011}. More recently, machine-learning approaches have been used for UAV motion planning in dynamic 3D space, showing that 3D UAV planning remains an active research problem \citep{Garibeh2025}.

Existing 3D UAV planning studies mostly search directly for collision-free paths or control policies within the UAV domain. Rather than searching a 3D path from scratch each time, this work transfers an obstacle-avoidance geometric structure learned on a different serial-chain platform as a structural prior for the UAV relay backbone.

\subsection{UAV Wireless Relay Networks}

Owing to their flexible deployment and high mobility, UAVs are useful in disaster response, temporary communication networks, and 5G/B5G auxiliary networks \citep{Zeng2016, Mozaffari2019}. UAV-based wireless networks must jointly consider 3D placement, trajectory optimization, a LoS-dominant channel, energy constraints, and QoS requirements, which gives them a design problem distinct from that of terrestrial networks \citep{Zeng2019, nikooroo2024optimization, chen2025multi}. In a relay backbone, overall performance is determined by the bottleneck capacity of the weakest hop and the accumulated end-to-end delay, so a geometrically collision-free path does not necessarily constitute a good communication backbone.

In multi-UAV systems, jointly determining trajectory, user association, and transmit power is a non-convex optimization problem \citep{8247211}, and the joint optimization of 3D trajectory and transmit power in UAV relay networks is likewise non-convex \citep{Zhang2018Relay}. These studies show that UAV relay placement is not merely path planning but a problem that couples geometry and wireless link quality.

Like prior UAV relay work, this work considers LoS, hop distance, bottleneck capacity, and end-to-end delay, but it does not search for the initial relay structure directly in the UAV domain. Instead, it transfers an ordered geometric skeleton learned in a robot-arm source domain and uses it as the starting point for communication-aware relay placement.

\subsection{Cross-Embodiment Robot Learning}

Knowledge transfer between different robot platforms is an important topic in robot learning. \citet{Devin2017} performed multi-task, multi-robot transfer through a modular neural network policy that combines task-specific and robot-specific modules. \citet{Chen2018} showed, through hardware-conditioned policies, that policy transfer is possible across robots with different kinematic structures, link lengths, and degrees of freedom. These studies show that robots with different physical structures can share common structural knowledge given an appropriate representation.

Existing cross-embodiment studies focus mainly on transferring control policies, locomotion, or manipulation skills between robots. This work extends that perspective to UAV relay networks. Although robot arms and UAV relay backbones differ in dynamics and control, they share a serial-chain structure in which intermediate nodes are ordered between a start anchor and a goal anchor. This work therefore does not transfer the robot arm's actuator-level commands or dynamics directly. Instead, it uses the ordered geometric skeleton obtained through forward kinematics as a structural prior for UAV relay placement.

\section{System Design}
\label{sec:system_design}

\subsection{Overview}

\begin{figure*}[t]
\centering
\includegraphics[width=0.8\textwidth, height=0.4\textwidth]{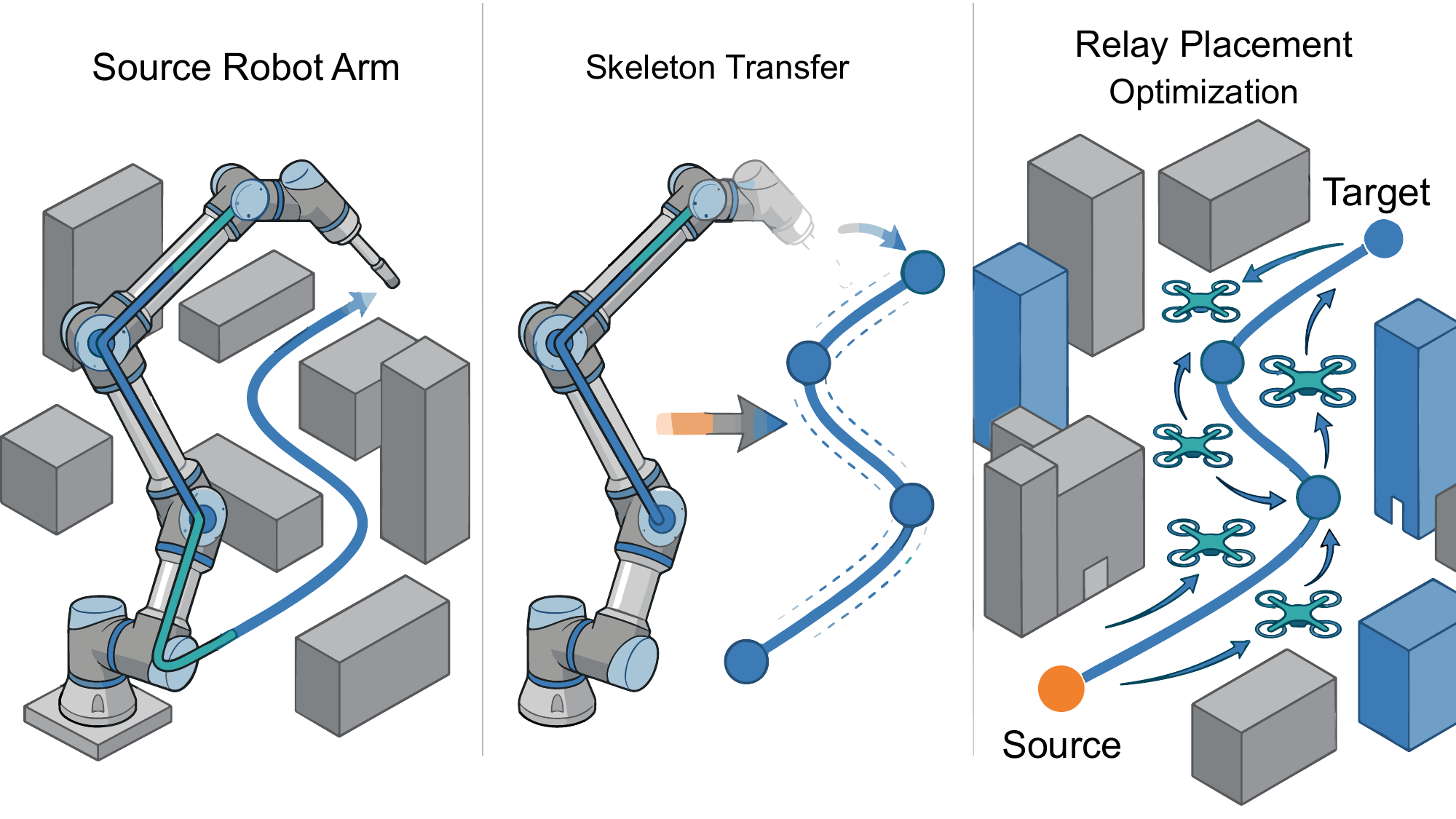}
\caption{Overview of the Arm2Air pipeline. A robot-arm skeleton is aligned to the gateway-target axis, a skeleton-conditioned transfer platform predicts the relay coordinates, and a communication-aware step refines them into the final relay backbone.}
\label{fig:system_overview}
\end{figure*}

Arm2Air transfers obstacle-avoidance geometric skeletons from heterogeneous robot arms to UAV relay backbone formation. Rather than searching for relay positions from scratch in a 3D urban environment, it reuses an ordered geometric prior learned on a different platform. The pipeline has four stages. First, the urban environment is encoded as a gateway, a target, building obstacles, an operable 3D region, and communication parameters. Second, robot-arm motions generated by a pretrained Neural MP model are converted through forward kinematics into ordered skeletons, which are aligned with the gateway-to-target axis and converted into initial relay chains. Third, a transformer-based transfer platform pretrained on these source skeletons predicts target-domain relay coordinates from the obstacle point cloud, global scene features, and the aligned source skeleton. Fourth, the predicted coordinates pass through a communication-aware refinement stage that enforces LoS, hop distance, safety distance, altitude, bottleneck capacity, delay, and movement cost. Fig.~\ref{fig:system_overview} summarizes the pipeline.

\subsection{Relay Scenario Representation}

A relay scenario is $\mathcal{E}=(g,t,\mathcal{B},\Omega,N)$ with gateway $g$, target $t$, building set $\mathcal{B}$, low-altitude workspace $\Omega\subset\mathbb{R}^3$, and relay count $N$. A placement is the ordered chain $X=[x_0,\dots,x_{N+1}]$ with $x_0=g$, $x_{N+1}=t$, and relays $x_i\in\mathbb{R}^3$, where consecutive nodes form one wireless hop. Let $d_i=\|x_i-x_{i+1}\|$ be the hop distance and $\ell_i\in\{0,1\}$ indicate whether hop $i$ keeps line of sight through $\mathcal{B}$. A hop is valid when $\ell_i=1$ and $d_i\le r_{\max}$. A blocked hop is unavailable and carries zero capacity, and a valid hop follows a distance-based Shannon model
\begin{equation}
c_i=\ell_i\,B\log_2\!\Big(1+\frac{P_t}{(N_0+I)\,d_i^{\alpha}}\Big),
\label{eq:capacity}
\end{equation}
with bandwidth $B$, transmit power $P_t$, noise power $N_0$, interference power $I$, and LoS path-loss exponent $\alpha$. For delay-based objectives, we use the clipped capacity $\bar c_i=\max(c_i,\epsilon_c)$ with a small constant $\epsilon_c>0$, so blocked or near-zero-capacity hops receive a large but finite delay penalty. The backbone throughput is bounded by the weakest hop, $C_{\mathrm{bot}}(X)=\min_i c_i$. A good placement keeps every hop valid while securing high $C_{\mathrm{bot}}$ and low delay.

\subsection{Source Geometric Skeleton Transfer}

The source domain is the 3D motion-planning problem of serial robot arms. An arm forms an ordered chain of joints from a start anchor to a goal anchor, and Arm2Air transfers the resulting sequence of 3D joint positions obtained through forward kinematics rather than joint angles or torques. Let $P=[p_0,\dots,p_{N+1}]$ be the source skeleton from base $p_0$ to end-effector $p_{N+1}$. It is obtained from the kinematic structure and joint configuration of the arm, and in the transfer stage it is used only as a spatial structure of intermediate nodes. We normalize the skeleton by its base-to-tip span $\|p_{N+1}-p_0\|$ into $\hat p_i$ and align it to the UAV gateway-target axis,
\begin{equation}
x_i^{0}=g+R_{\mathcal{E}}\,S_{\mathcal{E}}\,\hat p_i+h_{\mathcal{E}}(\hat p_i)\,e_z,
\label{eq:align}
\end{equation}
where $R_{\mathcal{E}}$ rotates the horizontal direction, $S_{\mathcal{E}}$ scales the formation to the city map, and $h_{\mathcal{E}}$ maps vertical variation into the low-altitude corridor. The aligned chain $X^0$ is the structural prior that detours around obstacles before any direct 3D search begins.

\subsection{Skeleton-Conditioned Relay Prediction}

The predictor is a transformer-based transfer platform over four token groups: a global scene token, obstacle point-cloud tokens, source-skeleton tokens carrying $\hat p_i$, and $N$ learned relay query tokens, each tagged with a type and positional embedding. A self-attention encoder mixes the concatenated tokens, and the query outputs pass through a bounded head that maps normalized coordinates into the workspace $\Omega$, yielding the target-domain relay prediction $\tilde X=[\tilde x_1,\dots,\tilde x_N]$. Training minimizes the coordinate error to a teacher placement $Y^{\ast}$,
\begin{equation}
\mathcal{L}=\frac{1}{N}\sum_{i=1}^{N}\|\tilde x_i-y_i^{\ast}\|^2,\qquad \mathrm{RMSE}=\sqrt{\mathcal{L}}.
\label{eq:loss}
\end{equation}
Training has two stages. Source pretraining uses ordered robot-arm skeletons derived from Neural MP motions as teachers, which makes the platform learn an ordered-chain prior. Target adaptation freezes the pretrained backbone and inserts low-rank adapters into the transformer feed-forward layers,
\begin{equation}
h=W_0\,u+\frac{\alpha}{r}\,B\,A\,u,
\label{eq:lora}
\end{equation}
with frozen $W_0$, trainable $A\in\mathbb{R}^{r\times d}$ and $B\in\mathbb{R}^{d\times r}$, rank $r\ll d$, and scale $\alpha/r$. Only the adapters, the relay query embeddings, and the output head are updated, so the platform predicts feasible teacher placements that reflect the LoS, distance, altitude, and capacity constraints of the UAV domain from limited target data while remaining conditioned on the aligned source skeleton.

\subsection{Communication-Aware Refinement}

The predicted chain $\tilde X$ is not used directly. Starting from $\tilde X$ and using the aligned source chain $X^0$ as a structural prior, Arm2Air minimizes a communication-aware objective over the workspace,
\begin{equation}
\begin{aligned}
J(X)=&-C_{\mathrm{bot}}(X)+\lambda_{\mathrm{los}}\Phi_{\mathrm{los}}(X)\\
&+\lambda_{d}\sum_i (d_i-r_{\max})_+^2+\lambda_{a}\sum_i \phi_a(z_i)\\
&+\lambda_{s}\sum_{i<j}\big(d_{\mathrm{safe}}-\|x_i-x_j\|\big)_+^2\\
&+\lambda_{m}\sum_i \|x_i-x_i^{0}\|^2+\lambda_{b}\,\Phi_{\mathrm{bnd}}(X)\\
&+\lambda_{\tau}\,D_{\mathrm{e2e}}(X),
\end{aligned}
\label{eq:objective}
\end{equation}
where $(\cdot)_+=\max(0,\cdot)$ and $\phi_a(z_i)=(z_{\min}-z_i)_+^2+(z_i-z_{\max})_+^2$ is the altitude term. The negative term $-C_{\mathrm{bot}}$ maximizes bottleneck capacity, $\Phi_{\mathrm{los}}$ and $\Phi_{\mathrm{bnd}}$ penalize building intersection and workspace exit, $D_{\mathrm{e2e}}(X)=\sum_i L/\bar c_i + (N+1)\tau$ is the clipped end-to-end delay, and the weights $\lambda$ are fixed constants. The refined chain minimizes $J$ over the workspace, $X^{\ast}=\arg\min_{X\in\Omega} J(X)$.

The refinement follows a feasibility-first strategy. Candidate chains are initialized from the transfer-platform prediction, the aligned source skeleton, interpolations between them, and geometrically diverse samples. Hard constraints are evaluated before the continuous objective, preventing a high-capacity but infeasible placement from replacing a valid backbone. When the initial solution does not provide sufficient communication quality, the transferred skeleton defines a restricted region for topology search and subsequent endpoint adjustment. This hierarchical procedure concentrates the search near obstacle-aware geometry while preserving the ability to explore alternative LoS corridors. All quality criteria and optimization settings are selected on the validation set and remain fixed during evaluation.

\subsection{Transfer Inference}

Algorithm~\ref{alg:arm2air} summarizes the full procedure. Source pretraining and LoRA adaptation are performed once. At inference time, Arm2Air queries a source skeleton, aligns it to the gateway-target axis, predicts target-domain relay coordinates conditioned on that aligned prior, and then refines the candidate chain.

\begin{algorithm}[t]
\caption{Arm2Air skeleton transfer and relay formation}
\label{alg:arm2air}
\begin{algorithmic}[1]
\REQUIRE source arm tasks, scenario $\mathcal{E}=(g,t,\mathcal{B},\Omega,N)$
\STATE Generate robot-arm motions with pretrained Neural MP and convert them into ordered skeletons
\STATE Pretrain transfer platform $f_\theta$ on the source skeletons \COMMENT{Eq.~(\ref{eq:loss})}
\STATE Freeze $\theta$ and add LoRA adapters to the FFN layers \COMMENT{Eq.~(\ref{eq:lora})}
\STATE Fine-tune adapters, query tokens, and output head on target teachers
\STATE Query a Neural MP-derived source skeleton $P$ for relay count $N$ and align it to obtain the prior $X^0$ \COMMENT{Eq.~(\ref{eq:align})}
\STATE $\tilde X \gets f_\theta(\mathcal{E},X^0)$ \COMMENT{skeleton-conditioned relay prediction}
\STATE Construct candidate chains from $\tilde X$, $X^0$, and geometrically diverse samples
\STATE Apply feasibility-first refinement under the communication objective
\STATE If communication quality is insufficient, search alternative topology near the transferred skeleton
\STATE $X^{\ast} \gets \arg\min_{X\in\Omega} J(X)$ among feasible candidates \COMMENT{Eq.~(\ref{eq:objective})}
\RETURN backbone $X^{\ast}$ with $C_{\mathrm{bot}}$, $D_{\mathrm{e2e}}$, and $\bar M$
\end{algorithmic}
\end{algorithm}

\subsection{Backbone Feasibility and Metrics}

A chain is feasible when every hop keeps LoS and range, each capacity meets the minimum rate $c_i\ge c_{\min}$, and all relays satisfy the safety distance, the altitude corridor, and the workspace bounds. We report the bottleneck capacity $C_{\mathrm{bot}}=\min_i c_i$ and the end-to-end delay $D_{\mathrm{e2e}}$, which sums the per-hop transmission delay $L/\bar c_i$ over packet size $L$ and the processing delay $\tau$ along the chain. We also report the mean relay displacement $\bar M=N^{-1}\sum_i\|x_i-x_i^{0}\|$, measured from the aligned source chain to the final placement. This metric is the interpretable counterpart of the squared movement term in Eq.~(\ref{eq:objective}) and quantifies how much adjustment is required after transfer. Effective bottleneck capacity equals $C_{\mathrm{bot}}$ for a hard-feasible chain and zero otherwise, ensuring that failed outputs remain visible. Delay is compared on the common subset for which every evaluated method returns a hard-feasible chain.

\subsection{Design Rationale}

Instead of solving the 3D relay-placement problem as a new path search each time, Arm2Air reuses an obstacle-avoidance structure already learned on a serial-chain platform. Robot arms and UAV relay networks differ in physical control but share the common structure of ordering intermediate nodes between a start and a goal. Arm2Air abstracts this transferable structure as an ordered geometric skeleton and then refines it with the communication constraints of the UAV domain. This design serves three purposes. It avoids low-level transfer failure, narrows the search through structured initialization, and refines each output using UAV communication metrics rather than geometry alone.

\section{Experiments}
\label{sec:experiments}

\subsection{Experimental Setup}

Each case contains a gateway, a target, and a procedurally generated 3D urban map represented by 96 obstacle tokens. Source skeletons are generated using the pretrained Neural MP model. Independent random seeds vary building position, footprint, and height, producing layouts that contain buildings of different scales. The target-domain data comprise 90 training maps, 30 validation maps, and nine planning-test maps. Communication is evaluated on a separate holdout of 30 newly seeded dense downtown maps with clustered buildings of varying positions, footprints, and heights. All partitions use disjoint random seeds, and neither test set is used for training or hyperparameter selection. The target-domain maps use a workspace scale of $3\times$ horizontally and $2\times$ vertically. The search-burden benchmark separately varies map scale from $1\times$ to $20\times$ on paired layouts with identical horizontal obstacle geometry.

The channel model assumes 10 MHz bandwidth, a LoS exponent of 2.1, a minimum rate of 2 Mbps, a 1 Mbit packet, and 2 ms processing delay per hop. Objective weights and optimization settings are selected on the validation split and fixed for all test maps. Communication range and safety distance scale with the horizontal workspace size. All experiments run on an Intel Core i5-12400 CPU and an NVIDIA RTX 3060 Ti GPU. We report median planning runtime, median path length, and mean communication and displacement metrics. Paired continuous outcomes are assessed using two-sided Wilcoxon signed-rank tests.

\subsection{Cost of Three-Dimensional Search}

This experiment tests whether 3D search cost grows enough to motivate transfer-based initialization. We paired 2D and 3D maps with identical horizontal obstacle layouts and increased map scale from $1\times$ to $20\times$. A*, Dijkstra, and D* used the same nine maps at each scale. Relay conversion and communication refinement were excluded.

\begin{figure}[t]
\centering
\includegraphics[width=0.9\columnwidth, height=1.3\columnwidth]{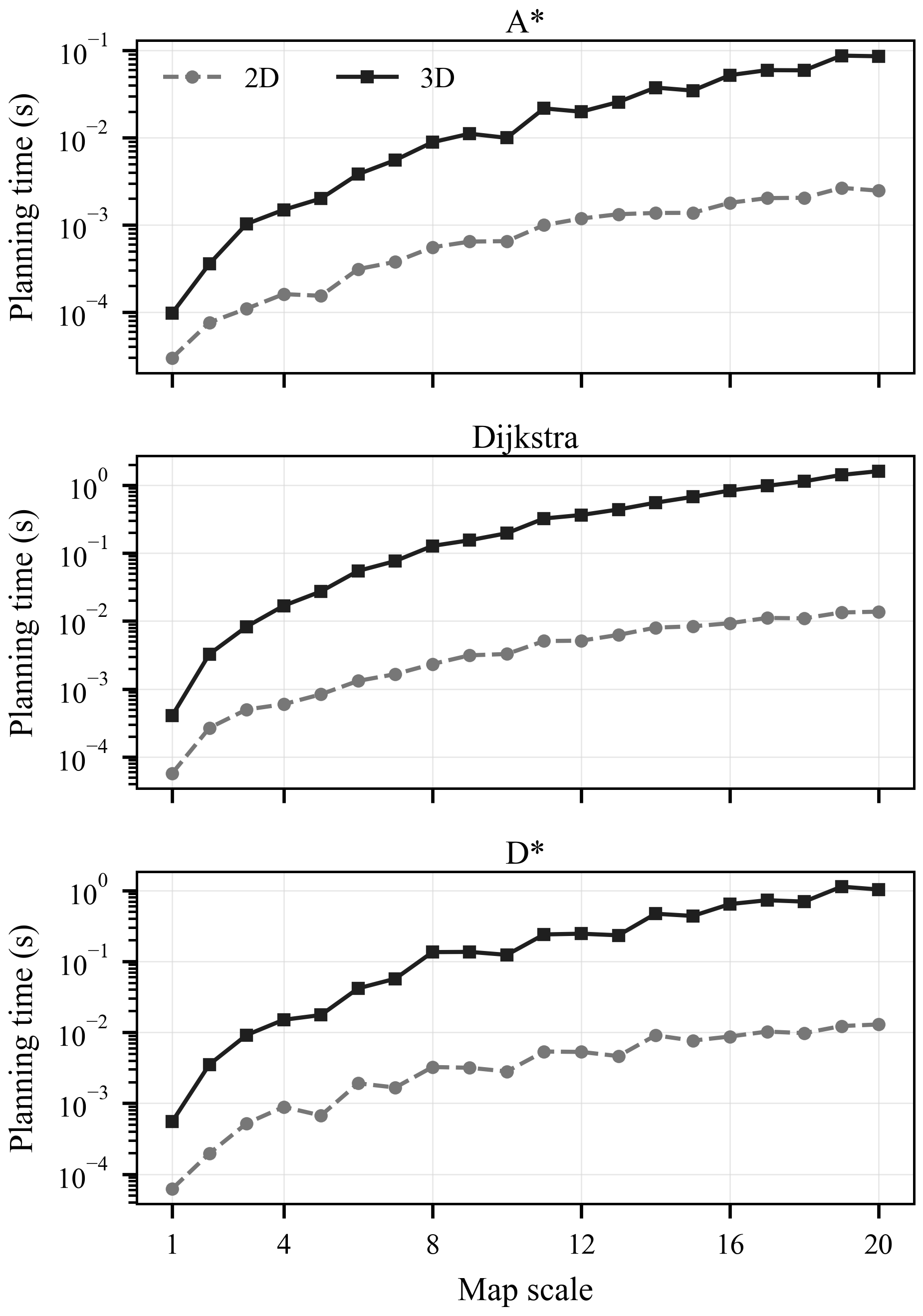}
\caption{Median planning time on paired 2D and 3D maps. The vertical axis is logarithmic.}
\label{fig:arm2air_2d3d}
\end{figure}

Fig.~\ref{fig:arm2air_2d3d} shows the resulting separation between paired 2D and 3D searches as the map grows. At $20\times$ scale, 3D planning took 34.8 times longer for A*, 79.9 times longer for D*, and 116.5 times longer for Dijkstra than the paired 2D searches. Explored states increased by factors of 10.2, 10.4, and 37.8. These results confirm that map scaling amplifies the computational penalty of 3D search and motivates transfer-based initialization.

\subsection{Planning Performance}

This experiment examines whether the transferred skeleton reduces the computation required to construct a complete relay backbone without sacrificing placement quality. A*, D*, Dijkstra, and RRT generate collision-free trajectories, which are resampled into seven ordered relay positions and processed under the same communication constraints as Arm2Air \citep{Karaman2011, Dijkstra1959}. Runtime covers the complete procedure required to produce the final seven-relay chain. All evaluated outputs satisfy the same LoS, range, rate, altitude, obstacle, and workspace constraints.

\begin{table}[t]
\centering
\scriptsize
\setlength{\tabcolsep}{1.5pt}
\begin{tabular*}{\columnwidth}{@{\extracolsep{\fill}}lcccc@{}}
\toprule
Method & Runtime & Length & $C_{\min}$ & Delay \\
 & s $\downarrow$ & m $\downarrow$ & Mbps $\uparrow$ & ms $\downarrow$ \\
\midrule
\textbf{Arm2Air} & \textbf{13.930} & \textbf{72.719} & \textbf{3.897} & \textbf{1994.44} \\
A* & 39.907 & 76.743 & 2.201 & 2317.93 \\
D* & 39.759 & 77.178 & 2.201 & 2287.61 \\
Dijkstra & 39.791 & 77.178 & 2.362 & 2310.59 \\
RRT & 39.741 & 84.400 & 2.080 & 2672.68 \\
\bottomrule
\end{tabular*}
\caption{End-to-end planning performance on nine high-clutter held-out maps. All values are medians.}
\label{tab:arm2air_planning_stress}
\end{table}

\begin{figure}[t]
\centering
\includegraphics[width=0.84\columnwidth, height=0.55\columnwidth]{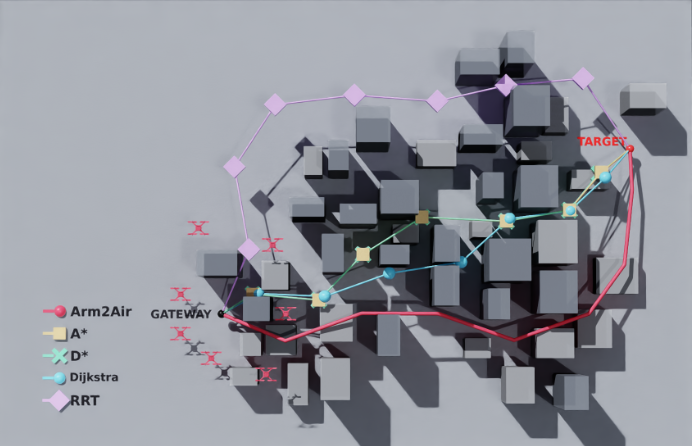}\\[2pt]
{\small (a) Initial}\\[4pt]
\includegraphics[width=0.84\columnwidth, height=0.55\columnwidth]{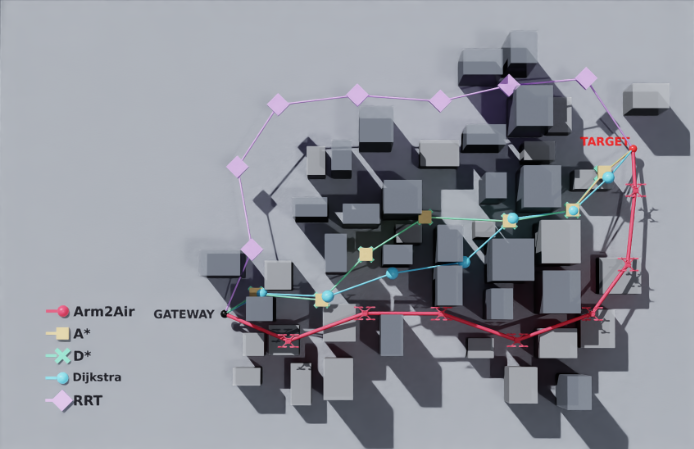}\\[2pt]
{\small (b) Final}
\caption{Relay deployment on a held-out dense urban map in Isaac Sim, comparing Arm2Air with conventional planners (A*, D*, Dijkstra, RRT). (a) Initial candidate paths with UAVs staged at the gateway; (b) final Arm2Air backbone routed around the buildings.}
\label{fig:arm2air_analogy}
\end{figure}

\begin{figure*}[t]
\centering
\includegraphics[width=0.850\textwidth, height=0.45\textwidth]{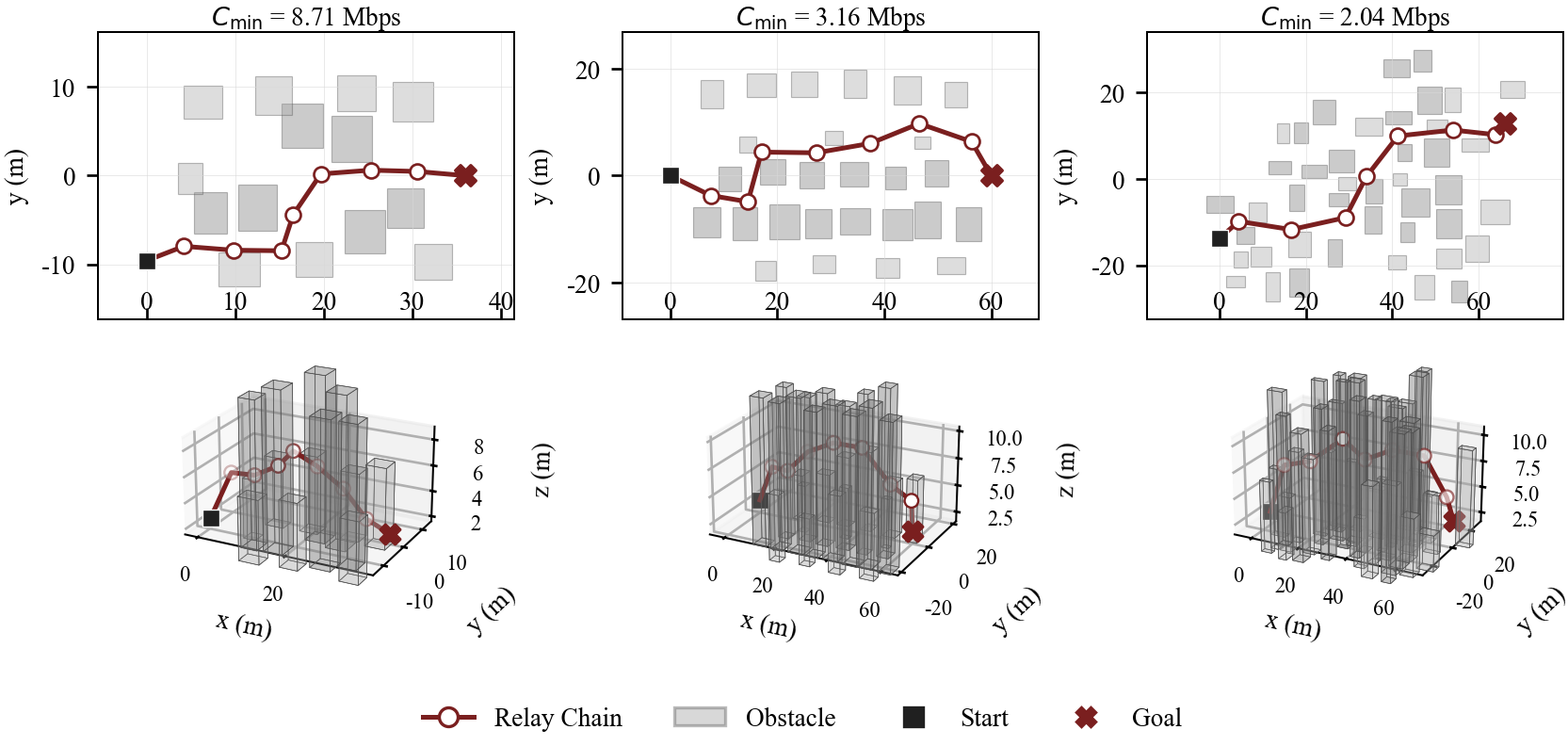}
\caption{Representative Arm2Air chains on held-out maps. The upper row shows planar projections and the lower row shows the same formations in 3D.}
\label{fig:arm2air_qualitative}
\end{figure*}

Table~\ref{tab:arm2air_planning_stress} shows that Arm2Air reduced median runtime by 64.9\% relative to RRT, the fastest conventional planner. It produced a 5.2\% shorter chain than A*, which achieved the shortest baseline result. Compared with the strongest baseline result for each communication metric, Arm2Air increased bottleneck capacity by 65.0\% and reduced end-to-end delay by 12.8\%. Arm2Air achieved better length, capacity, and delay on all nine maps. These results show that the transferred skeleton reduces search cost while providing an initialization suited to serial relay formation. Fig.~\ref{fig:arm2air_analogy} contrasts the candidate paths of all planners with the final Arm2Air deployment on a representative Isaac Sim map, and Fig.~\ref{fig:arm2air_qualitative} presents representative planar and 3D relay formations.

\subsection{Communication Performance}

We compare Arm2Air with IMPC-MD, which combines MiniEdgeRRT* and distributed model predictive control for obstacle-aware LoS deployment \citep{chen2025multi}, and AO Placement, which applies alternating optimization to 3D relay placement and communication-resource allocation \citep{nikooroo2024optimization}. All methods use seven relays under identical evaluation conditions. We evaluate the eight hops using bottleneck capacity $\cmin$, capacity variance $\sigma_C^2$, maximum hop distance $D_{\max}$, hop-distance variance $\sigma_D^2$, and Move. The variances measure capacity and distance balance, while Move denotes the mean 3D displacement from the aligned source chain. Infeasible chains are assigned $\cmin=0$. Higher $\cmin$ and lower values for the remaining metrics are preferred.

\begin{table}[t]
\centering
\scriptsize
\setlength{\tabcolsep}{1.0pt}
\begin{tabular*}{\columnwidth}{@{\extracolsep{\fill}}lrrrrr@{}}
\toprule
Method & $\cmin$ & $\sigma_C^2$ & $D_{\max}$ & $\sigma_D^2$ & Move \\
 & Mbps $\uparrow$ & Mbps$^2$ $\downarrow$ & m $\downarrow$ & m$^2$ $\downarrow$ & m $\downarrow$ \\
\midrule
\textbf{Arm2Air}
& \textbf{3.772}
& \textbf{0.147}
& \textbf{9.463}
& \textbf{0.190}
& \textbf{14.639} \\
IMPC-MD
& 2.844
& 0.584
& 10.905
& 0.767
& 17.612 \\
AO Placement
& 2.061
& 281.484
& 12.976
& 21.852
& 15.714 \\
\bottomrule
\end{tabular*}
\caption{Communication quality and hop balance on nine high-obstruction urban maps.}
\label{tab:arm2air_communication}
\end{table}

As shown in Table~\ref{tab:arm2air_communication}, Arm2Air achieved the best mean result for all five metrics. Compared with IMPC-MD, Arm2Air increased $\cmin$ by 32.6\%, reduced capacity variance by 74.7\%, reduced $D_{\max}$ by 13.2\%, reduced hop-distance variance by 75.2\%, and reduced Move by 16.9\%. The improvements in $\cmin$ and $D_{\max}$ were statistically significant with $p=0.006$. These results indicate that Arm2Air constructs stronger weakest links while distributing relay UAVs more evenly across the communication chain.

Compared with AO Placement, Arm2Air increased $\cmin$ by 83.1\%, reduced capacity variance by 99.9\%, reduced $D_{\max}$ by 27.1\%, reduced hop-distance variance by 99.1\%, and reduced Move by 6.8\%. The first four improvements were statistically significant with $p\leq0.002$. AO Placement occasionally placed relays close together, producing strong local links while leaving a long and weak bottleneck hop. In contrast, the transferred skeleton provides an ordered initialization that distributes the relays along the available LoS corridor before communication-aware refinement.

\subsection{Movement Efficiency}

This experiment tests whether the transferred skeleton reduces the geometric adjustment required to obtain a communication-feasible formation. The Move column in Table~\ref{tab:arm2air_communication} reports the mean 3D displacement of each relay from the aligned source chain to the final placement. Arm2Air required 14.639 m, compared with 17.612 m for IMPC-MD and 15.714 m for AO Placement. These values correspond to mean reductions of 16.9\% and 6.8\%, respectively.

Arm2Air remained the lowest-displacement method while providing the highest bottleneck capacity. Although the paired movement difference was not significant, the lower mean adjustment accompanies significant gains in communication quality and hop balance. This result supports the transferred skeleton as a geometrically informed prior rather than an optimization shortcut alone.

\subsection{Ablation Study}

\textbf{Data-efficient target adaptation.}
We test whether source pretraining improves relay prediction when only a small amount of target-domain data is available. One training map is selected from each of three map families, giving three target maps per subset. Five independently sampled subsets are evaluated on the same 30 held-out target-domain maps. Scratch uses random initialization, Full Fine-tuning updates the source-pretrained model, and Arm2Air performs LoRA-based target adaptation.

\begin{table}[t]
\centering
\footnotesize
\setlength{\tabcolsep}{4.0pt}
\begin{tabular*}{\columnwidth}{@{\extracolsep{\fill}}lcc@{}}
\toprule
Method & RMSE & Trainable Parameters \\
 & m $\downarrow$ & \\
\midrule
Scratch & 6.669 $\pm$ 1.409 & 1.383M \\
Full Fine-tuning & 3.241 $\pm$ 0.099 & 1.383M \\
\textbf{Arm2Air} & \textbf{3.092 $\pm$ 0.146} & \textbf{0.134M} \\
\bottomrule
\end{tabular*}
\caption{Relay-position RMSE using one training map per map family. Values are mean $\pm$ standard deviation over five target-subset seeds.}
\label{tab:arm2air_oneshot}
\end{table}

Table~\ref{tab:arm2air_oneshot} shows that Arm2Air reduced RMSE by 53.6\% relative to Scratch and achieved lower error on all five target-subset seeds. Arm2Air also obtained a 4.6\% lower mean RMSE than Full Fine-tuning while updating 0.134 million parameters instead of 1.383 million. This difference from Full Fine-tuning was not statistically significant with five subsets. The transferred representation enables accurate adaptation using only three maps and far fewer parameters.

\textbf{Input and training-stage ablation.}
We evaluate the direct Transformer output before communication-aware refinement so that post-processing cannot compensate for missing information. No Source Prior masks the aligned skeleton tokens. No PCD masks the obstacle point-cloud tokens. No LoRA uses the source-pretrained checkpoint before target adaptation. No Communication FT uses the pre-communication LoRA checkpoint. All variants share the same 30 test maps.

\begin{table}[t]
\centering
\footnotesize
\setlength{\tabcolsep}{2.4pt}
\begin{tabular*}{\columnwidth}{@{\extracolsep{\fill}}lcc@{}}
\toprule
Variant & RMSE & Degradation \\
 & m $\downarrow$ & \% \\
\midrule
\textbf{Arm2Air} & \textbf{3.156 $\pm$ 1.325} & -- \\
No Source Prior & 3.704 $\pm$ 1.636 & 17.4 \\
No PCD & 8.071 $\pm$ 2.112 & 155.7 \\
No LoRA & 7.407 $\pm$ 1.975 & 134.7 \\
No Communication FT & 7.922 $\pm$ 1.784 & 151.0 \\
\bottomrule
\end{tabular*}
\caption{Direct relay-position error before communication refinement.}
\label{tab:arm2air_direct_ablation}
\end{table}

Table~\ref{tab:arm2air_direct_ablation} shows that complete Arm2Air achieved the lowest relay-position error. Removing the source prior increased RMSE by 17.4\%. This moderate degradation occurs because the point cloud, global scene features, and relay query tokens still provide sufficient information to reconstruct a coarse relay chain. The transferred skeleton therefore acts as a geometric bias that improves the initial chain structure without replacing the target-domain obstacle observation. Removing the point cloud increased RMSE by 155.7\%, showing that the model relies on the observed obstacle geometry rather than reproducing a fixed chain. The models without LoRA and communication-oriented fine-tuning increased RMSE by 134.7\% and 151.0\%, respectively. Arm2Air outperformed the four variants on 24, 30, 29, and 29 of the 30 paired maps. Scene observations capture obstacles, while the source skeleton and target adaptation improve initial relay placement.



\section{Conclusion}
\label{sec:conclusion}

Arm2Air transfers ordered obstacle-avoidance skeletons derived from Neural MP robot-arm motions to constrained 3D UAV relay formation. Its transformer-based transfer platform models the source skeleton geometry, LoRA adapts the representation from limited UAV data, and communication-aware refinement optimizes LoS, range, capacity, delay, and movement without transferring joint commands or dynamics.

On nine held-out high-clutter 3D maps, Arm2Air reduced median end-to-end planning runtime by 64.9\% relative to the fastest conventional planner. In the geometry-defined high-obstruction group of the separate 30-map communication holdout, Arm2Air increased bottleneck capacity by 32.6\%, reduced capacity variance by 74.7\%, reduced maximum hop distance by 13.2\%, reduced hop-distance variance by 75.2\%, and reduced relay displacement by 16.9\% relative to IMPC-MD. With only three target-domain training maps, Arm2Air reduced relay-position RMSE by 53.6\% relative to Scratch while updating 0.134 million parameters, compared with 1.383 million for Scratch and Full Fine-tuning. The direct-output ablation further showed that the source skeleton improves the initial chain geometry, while the point cloud and target-domain adaptation provide the scene-specific information required for accurate relay placement.

These results show that cross-embodiment skeleton transfer efficiently initializes communication-constrained UAV relay formation. By transferring ordered relations rather than UAV-specific controls, Arm2Air may extend to other chain-structured tasks, including multi-robot formation and sensor deployment. Future work will examine this transferability under online point clouds, dynamic obstacles, channel uncertainty, and physical multi-UAV deployment.

\bibliography{references}

\end{document}